\documentclass[letterpaper, 10 pt, conference]{ieeeconf}  

\IEEEoverridecommandlockouts                              

\usepackage{amsmath}
\usepackage{amssymb}
\usepackage{graphicx}
\usepackage{booktabs}
\usepackage{multirow}
\usepackage{caption}
\usepackage[table]{xcolor}
\PassOptionsToPackage{hyphens}{url}
\usepackage{url}
\definecolor{myemerald}{rgb}{0.753, 0.898, 0.804}
\definecolor{mylightgreen}{rgb}{0.894, 0.933, 0.745}
\definecolor{myyellow}{rgb}{0.996, 0.972, 0.780}
\newcommand{\firstc}{\cellcolor{myemerald!100}}    
\newcommand{\secondc}{\cellcolor{mylightgreen!100}} 
\usepackage{tikz}
\usetikzlibrary{arrows.meta, calc, positioning}
\newcommand{\method}{RFL} 

\title{\LARGE \bf
Rendering-Free Lookahead for Question-Guided Active Vision
}

\author{Koya Sakamoto$^{1}$ \and
Daichi Azuma$^{1}$ \and
Shuhei Kurita$^{2,3,4}$ \and
Naoya Chiba$^{5}$ \and[\hfill\mbox{}\newline\mbox{}\hfill]
Yusuke Iwasawa$^{1}$ \and
Yutaka Matsuo$^{1}$ \and
Taiki Miyanishi$^{1}$
\thanks{$^{1}$The University of Tokyo, Japan}%
\thanks{$^{2}$National Institute of Informatics, Japan}%
\thanks{$^{3}$Institute of Science Tokyo, Japan}%
\thanks{$^{4}$NII LLMC, Japan}%
\thanks{$^{5}$The University of Osaka, Japan}%
}

\begin{document}

\maketitle
\thispagestyle{empty}
\pagestyle{empty}

 \begin{abstract}
Active robot vision requires controlling the camera to reveal task-relevant information that is hidden from the current viewpoint. For example, determining what is inside a box may require raising the camera and looking down into it. For viewpoint-dependent question answering, the challenge is to select camera motions that expose the visual evidence needed to answer the question.
Although vision-language models (VLMs) can interpret observed images, selecting such motions requires anticipating the usefulness of unseen views. 
We quantify this usefulness as \emph{answerability}, a VLM's estimate that a view suffices to answer the question, and present Rendering-Free Lookahead (RFL), a viewpoint-selection policy that ranks candidate camera motions by predicted future answerability. RFL transfers visual lookahead from deployment to offline training. At training, a privileged teacher renders candidate future views in 3D Gaussian Splatting (3DGS) scenes and uses a frozen VLM to compute one- and two-step answerability targets. Through two-stage distillation, a student learns to predict these action values from the question, recent visual observations, and a candidate camera motion.
At deployment, RFL uses these predicted values to select camera motions without rendering future views.
On 377 E3VS-Bench test episodes in unseen environments, RFL improves the mean judge score by 43\% over a direct-action baseline using the same VLM. These results support learning camera-control policies from privileged visual lookahead for viewpoint-dependent question answering.

\end{abstract}

\section{INTRODUCTION}
Visual perception in robotics depends on the actions used to acquire observations.
When task-relevant evidence is occluded, a robot must move its camera to reveal it.
For example, a mug is readily recognized from the side, yet determining whether it is empty may require raising the camera and looking into it.
Beyond locating the object, selecting such camera motions requires inferring where the evidence is likely to be and which viewpoint will reveal it.

We study this problem in Embodied 3D Visual Search (E3VS)~\cite{sakamoto2026e3vsbench}.
Given a question and egocentric RGB observations, an agent controls its camera position, yaw, and pitch in an unseen scene to reach a viewpoint from which the question can be answered.
Without a target viewpoint or scene map, the agent must infer from its observations which camera motion is likely to reveal the missing evidence within a limited action budget.
We consider the setting where arbitrary viewpoints can be rendered in the training scenes, whereas the deployment scene is unknown and only the agent's own observations are available.

\begin{figure}[t]
    \centering
    \includegraphics[width=\linewidth]{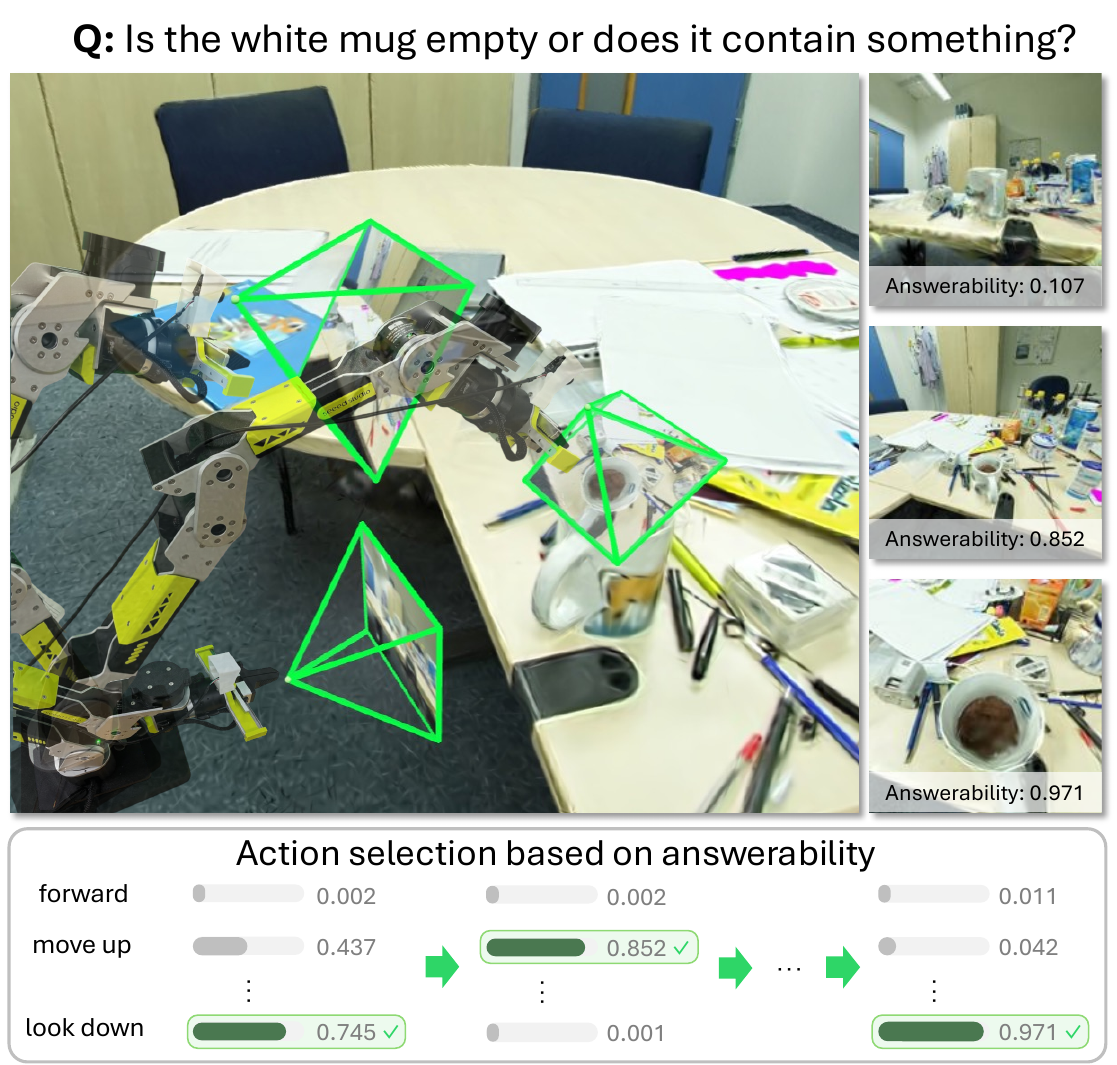}
    \vspace{-1.5em}
    \caption{\textbf{Rendering-free lookahead.} \method{} selects camera motions by predicted answerability, which rises as the agent closes in on a view that reveals the answer (right), with no rendering or generation at deployment.}
    \label{fig:teaser}
    \vspace{-1.65em}
\end{figure}

Vision-language models (VLMs) offer a foundation for this task by relating visual observations to natural-language questions~\cite{mahumar2024openeqa,koo2025towardb,wang2026planning}.
A central challenge in selecting camera motions is to anticipate whether an unseen future view will reveal the evidence needed for an answer.
Supervision from demonstrated actions alone specifies which motion to select but does not quantify how useful each alternative would be.
This motivates learning a question-conditioned value for each candidate motion from its counterfactual visual outcome during training.

We introduce Rendering-Free Lookahead (\method{}; Fig.~\ref{fig:teaser}), which ranks candidate camera motions by their predicted usefulness for answering the question.
We quantify the usefulness of a view through \emph{answerability}, a VLM's estimate of whether the view contains sufficient visual evidence to answer the question.
Prior work has used confidence to guide stopping~\cite{ren2024explore} and estimated answerability across scene locations~\cite{azuma2024answerability}.
In contrast, our key idea is to evaluate answerability at candidate future viewpoints during training and distill the resulting short-horizon action values into a camera-control policy.

To this end, a privileged teacher scores future views rendered from 3D Gaussian Splatting (3DGS) training scenes~\cite{kerbl2023GaussianSplatting} with a frozen VLM, constructing lookahead targets for every candidate motion, including unexecuted alternatives.
The student learns to predict these values from the question and recent observations, first on one-step and then on two-step targets, enabling viewpoint selection without rendering at deployment.

We evaluate \method{} on 377 E3VS-Bench test episodes across 21 unseen scenes.
It improves the mean judge score from 2.14 to 3.05 (+43\%) over the same VLM acting directly and also surpasses the strongest evaluated proprietary baseline, GPT-5.1 (2.81).
It also outperforms cloning the teacher's actions with the same backbone and trajectories (2.47), supporting candidate-wise value supervision for viewpoint selection.
Its one-step variant matches the mean judge score of the privileged one-step rendering oracle (2.98).

Our contributions are:
\begin{itemize}
\item \method{}, which calibrates VLM answerability into an action value by distilling privileged lookahead targets, requiring no rendering at deployment.
\item Improvements on unseen scenes over the same VLM acting directly, GPT-5.1, and cloning the teacher's actions.
\item Transfer of the simulation-trained policy to a real 6-DoF arm with an end-effector camera, achieving the highest mean judge score among the compared methods without test-time rendering.
\end{itemize}

\section{Related Work}

\subsection{Active Vision and Embodied Question Answering}
Active vision selects sensing actions to acquire task-relevant information~\cite{Bajcsy1988ActivePerception,Aloimonos1988ActiveVision}.
This principle underlies both learned gaze control for robotic manipulation~\cite{kerrj2025eyerobot} and question-driven exploration in active Embodied Question Answering (EQA)~\cite{das2018embodied,mahumar2024openeqa}.
In active EQA, a natural-language question defines the information-gathering objective, and agents explore to acquire the visual evidence needed for an answer, often using structured scene representations~\cite{grapheqa2025,sakamoto2024mapeqa}.
Closest to our use of answerability, Explore until Confident uses calibrated VLM confidence to decide when to stop exploring~\cite{ren2024explore}, and Answerability Fields estimates which locations in a scene are answerable~\cite{azuma2024answerability}.
Neither uses answerability to select camera motions, whereas \method{} predicts the answerability of candidate future viewpoints and uses it as the action value that drives viewpoint selection.

Closer to camera control, VG-AVS learns viewpoint adjustments through supervised fine-tuning and reinforcement learning with an answer-correctness reward~\cite{koo2025towardb}.
Planning with the Views uses self-exploration and view-graph distillation for target-view localization, given a target image and top-down reference~\cite{wang2026planning}.
\method{} operates in the E3VS setting, where neither a scene map nor a target viewpoint is supplied~\cite{sakamoto2026e3vsbench}.
While VG-AVS learns to output a viewpoint adjustment, \method{} learns to score candidate motions using graded, question-conditioned answerability targets over a short horizon.
This supervision covers every candidate motion at each training state, including alternatives not executed by the teacher.

\subsection{Visual Lookahead for Action Selection}

Anticipating the visual consequences of a motion provides a basis for selecting informative viewpoints.
In navigation, world models capture relationships between motions and future observations~\cite{koh2021pathdreamer,bar2025navigation,azuma2026navwam}, while semantic representations support lookahead at candidate viewpoints~\cite{wang2024lookahead}.
Additional views can also be rendered from persistent 3DGS scene memory to recover visual evidence~\cite{lu2026gsmem}.

Future prediction can also support policy learning without being required at deployment, through video co-training~\cite{yuan2026fastwam} or trajectory supervision derived from generated future views~\cite{chen2026worldmap}.
\method{} similarly uses future views during training, yet transfers their utility into question-conditioned action values based on short-horizon answerability.
These values are supervised using rendered future views and predicted directly by the deployed policy, without rendering or generating future observations.

\subsection{Privileged Supervision and Value Distillation}

The usefulness of a camera motion can be difficult to assess from current observations alone.
Privileged supervision exploits richer scene or state information during training to guide policies that operate from limited observations at deployment~\cite{chen2019learning,lee2020learning,ramakrishnan2018sidekick}.
Teacher supervision can extend beyond selected actions to action preferences and search-derived values~\cite{rusu2016policy,ruoss2024amortized}.
Action-value learning has also been explored for language and vision-language models through offline reinforcement learning~\cite{hong2025qsft,bai2025digiq}.

\method{} evaluates rendered future views with a frozen VLM and uses short-horizon search to construct graded, question-conditioned answerability targets.
These targets supervise every candidate camera motion at each sampled training state, including unexecuted alternatives.
The student thus learns to compare candidate motions by their usefulness for acquiring the visual evidence needed for an answer.
\section{Problem Statement}
\label{sec:problem}

We consider the Embodied 3D Visual Search (E3VS) task~\cite{sakamoto2026e3vsbench}, in which an agent controls its camera in an unseen scene to acquire the visual evidence needed to answer a question $q$.
The agent receives only $q$ and egocentric RGB observations, with neither a scene map nor a target viewpoint.

Let $S$ denote a 3D scene and $p_t$ the camera pose at step $t$, comprising a 3D position, yaw, and pitch with fixed roll.
The RGB observation is $o_t = R_S(p_t)$, where $R_S$ maps a camera pose to its observation in scene $S$.
The action space $\mathcal{A}$ consists of ten discrete camera motions (six translations and four rotations) and \texttt{stop}.
For an action $a_t \in \mathcal{A}$, the state transition is $p_{t+1} = T(p_t, a_t)$.
A colliding motion leaves the pose unchanged while consuming a step.

The episode terminates at step $\tau$ when the agent selects \texttt{stop} or reaches the action budget $T_{\max}$.
A VQA model produces $\hat{y} = \mathrm{VQA}(o_\tau, q)$ from the final observation, which is evaluated against the ground truth $y^{\ast}$.

\begin{figure*}[t]
    \centering
    \includegraphics[width=0.95\linewidth]{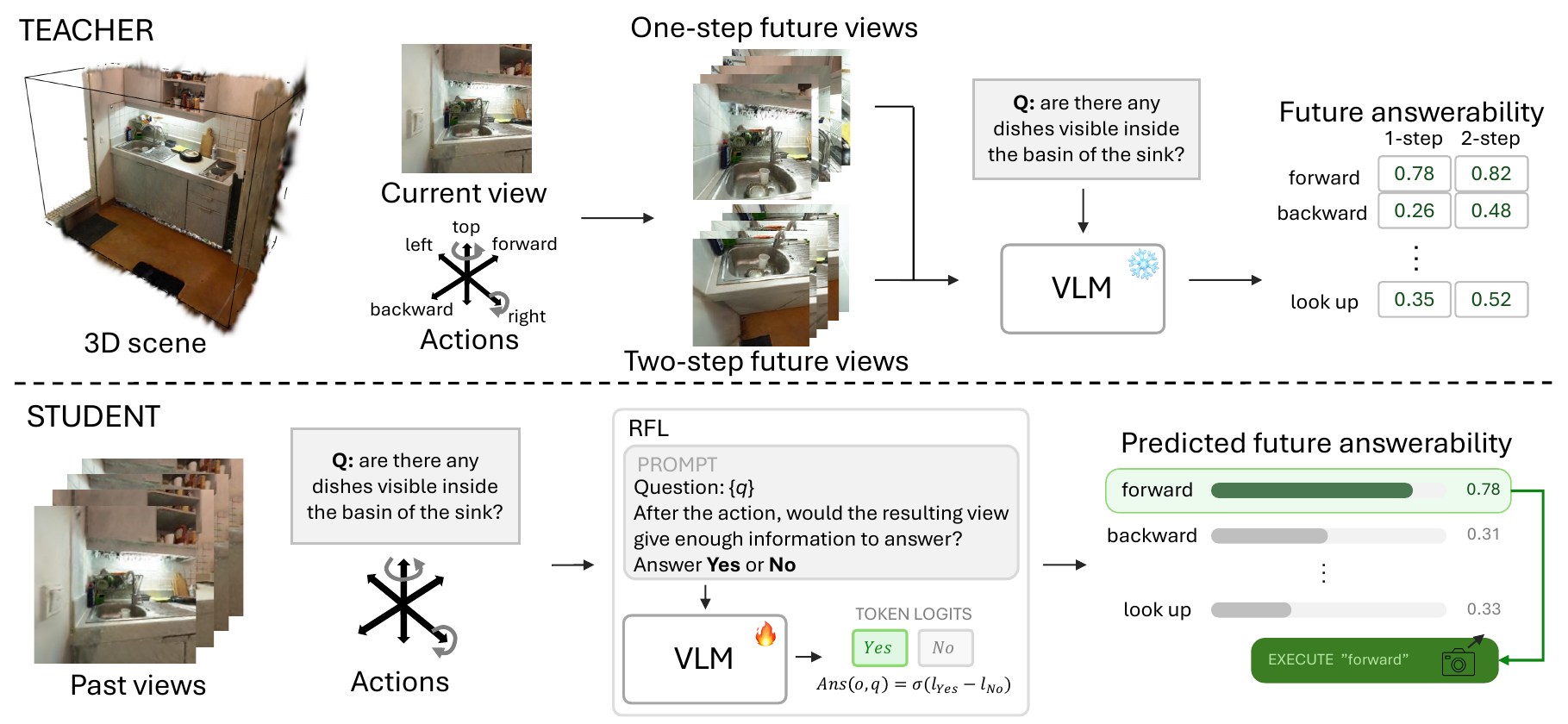}
    \vspace{-0.5em}
    \caption{\textbf{Overview of \method{}.}
    A privileged teacher renders each candidate action's one- and two-step future views in 3DGS and scores them with the frozen VLM's answerability probe, yielding the lookahead targets $Q_1$ and $Q_2$; the student learns to predict these values from the question, past views, and the action, requiring no rendering at deployment.}
    \label{fig:method}
    \vspace{-1.5em}
\end{figure*}

\section{Method}
\label{sec:method}

\method{} learns short-horizon action values from privileged visual lookahead, enabling camera control without rendering future views at deployment (Fig.~\ref{fig:method}).
An offline teacher renders future views in 3DGS scenes and evaluates them with a frozen VLM to construct answerability-based targets for all candidate camera motions (Sec.~\ref{sec:teacher}).
Through two-stage distillation, the student learns to predict these values from the question, recent observations, and a candidate motion expressed as text (Sec.~\ref{sec:distill}).
At deployment, the policy scores candidate motions, executes the highest-scoring one, and repeats with new observations (Sec.~\ref{sec:policy}).

\subsection{Answerability Probe}
To construct action-value targets, the teacher must assess whether candidate future views contain sufficient visual evidence to answer the question, which we estimate with a frozen VLM.
Given an observation $o$ and question $q$, we prompt the VLM with ``can $q$ be answered from this view?'' and introduce answerability $\mathrm{Ans}(\cdot)$ as
\begin{equation}
\mathrm{Ans}(o, q)
= \frac{e^{\ell_{\mathrm{Yes}}}}
{e^{\ell_{\mathrm{Yes}}} + e^{\ell_{\mathrm{No}}}}
= \sigma\big(\ell_{\mathrm{Yes}} - \ell_{\mathrm{No}}\big)
\in [0,1],
\label{eq:probe}
\end{equation}
where $\sigma$ is the sigmoid function.
$\ell_{\mathrm{Yes}}$ and $\ell_{\mathrm{No}}$ are obtained by applying logsumexp to the logits of each class's surface variants at the answer position.

\begin{figure}[t]
    \centering
    \includegraphics[width=0.8\linewidth]{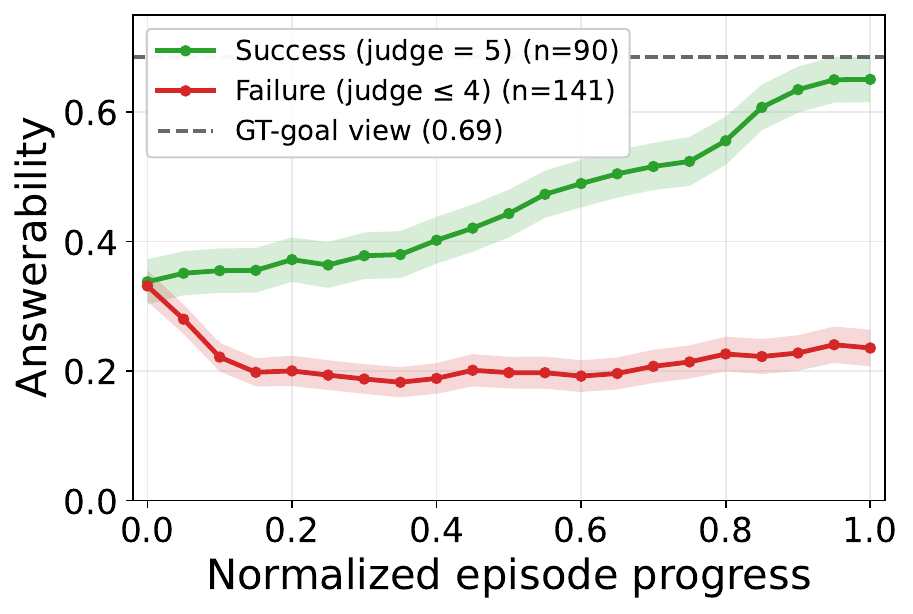}
    \vspace{-0.8em}
    \caption{\textbf{Answerability tracks viewpoint search.} Mean probed answerability vs.\ normalized progress on Gemini 3.0 Flash rollouts over the 231 validation episodes (bands: $\pm 1$\,SE; dotted: mean at annotated goal views; Sec.~\ref{sec:exp-signal}).}
    \label{fig:prelim}
    \vspace{-1.3em}
\end{figure}

On baseline rollouts, mean answerability increases during successful searches, whereas failed searches decline early and stay low (Fig.~\ref{fig:prelim}), motivating the probe as a source of action-value targets (Sec.~\ref{sec:exp-signal}).
The teacher applies the probe to rendered future views during short-horizon search.

\subsection{Privileged Lookahead Teacher}
\label{sec:teacher}

The teacher uses a 3DGS representation of each training scene to evaluate the visual consequences of candidate camera motions.
Let $s_t = (S,p_t,q)$ denote its privileged state and $\mathcal{A}_{\mathrm{cam}} = \mathcal{A} \setminus \{\texttt{stop}\}$ the set of camera motions.
At each visited state, the teacher constructs one-step and two-step lookahead targets, $Q_1(q,p_t,a)$ and $Q_2(q,p_t,a)$, for every candidate motion; their dependence on the scene $S$ is left implicit.

Let $o' = R_S(T(p_t,a))$ denote the view rendered after candidate motion $a$, and $o''$ the view after a further motion $a'$ from the resulting pose.
The one-step target is the answerability of the resulting view:
\begin{equation}
Q_1(q,p_t,a) = \mathrm{Ans}(o',q).
\label{eq:g1}
\end{equation}

A camera motion may reveal little evidence immediately yet enable a more informative subsequent view, motivating two-step lookahead.
To limit rendering, we expand only the top-$k$ ($k=3$) first motions ranked by $Q_1$, denoted by $\mathcal{B}_t$, a beam-style approximation that concentrates rendering on the most promising motions.
For each expanded motion, the two-step target is the highest answerability among its allowed continuations:
\begin{equation}
Q_2(q,p_t,a) =
\max_{\substack{a' \in \mathcal{A}_{\mathrm{cam}}\\
                a' \neq \mathrm{inv}(a)}}
\mathrm{Ans}(o'',q),
\qquad a \in \mathcal{B}_t.
\label{eq:g2}
\end{equation}
For the remaining candidates, we set $Q_2(q,p_t,a)=Q_1(q,p_t,a)$.
Here $\mathrm{inv}(a)$ is the inverse motion, excluded to prevent immediately undoing the first motion.
Since a second motion is required for expanded candidates, we retain $Q_2(q,p_t,a)$ even when it is lower than $Q_1(q,p_t,a)$.

The teacher generates training trajectories using $Q_2$ for motion selection and records both targets for every candidate motion at each visited state, including unexecuted alternatives.
All rendering and target generation are performed offline.

\subsection{Two-Stage Value Distillation}
\label{sec:distill}

The student learns to predict the teacher's lookahead values using only inputs available at deployment.
It is initialized from the same base VLM as the teacher's frozen probe and fine-tuned with LoRA adapters with trainable parameters $\theta$.
The teacher targets $Q_h(q,p_t,a)$ depend on the privileged state through rendering, whereas the student receives only the question $q$, a short sequence of recent RGB observations $H_t$, and a textual description of the candidate motion $a$, so that $H_t$ is used in place of $p_t$.
The student is asked whether the viewpoint reached after executing $a$ would provide enough information to answer $q$, and its action value is read out from the Yes/No logits as in Eq.~(\ref{eq:probe}):
\begin{equation}
\hat{Q}_\theta(q,H_t,a)
= \sigma\big(\ell^{\theta}_{\mathrm{Yes}}(q,H_t,a) - \ell^{\theta}_{\mathrm{No}}(q,H_t,a)\big),
\label{eq:student}
\end{equation}
where $\ell^{\theta}_{\mathrm{Yes}}$ and $\ell^{\theta}_{\mathrm{No}}$ are the logits of the adapted VLM at the answer position.
Whereas the teacher applies the readout of Eq.~(\ref{eq:probe}) to a rendered future view (Eqs.~(\ref{eq:g1}) and (\ref{eq:g2})), the student applies it to the current observations and the motion text, and distillation trains $\hat{Q}_\theta(q,H_t,a)$ to match $Q_h(q,p_t,a)$ at the states visited by the teacher.

The offline training dataset $\mathcal{D}$ contains records $(q,H_t,a,v_1,v_2)$ collected at visited training states, where $v_h = Q_h(q,p_t,a)$ is the teacher target at that state.
For each training stage $h \in \{1,2\}$, we fit the continuous targets with soft-label binary cross-entropy over $\mathcal{D}$:
\begin{equation}
\begin{aligned}
\mathcal{L}_h(\theta)
= -\mathbb{E}_{\mathcal{D}}\Big[
&v_h \log \hat{Q}_\theta(q,H_t,a)\\
&+ (1-v_h)\log\big(1-\hat{Q}_\theta(q,H_t,a)\big)
\Big].
\end{aligned}
\label{eq:loss}
\end{equation}

We first minimize $\mathcal{L}_1$ to learn one-step action values and then continue from the resulting parameters by minimizing $\mathcal{L}_2$ on the two-step targets, both on the offline teacher dataset.
Alternative observation histories are compared in Sec.~\ref{sec:ablation}.

\subsection{Inference-Time Policy}
\label{sec:policy}

At deployment, the policy selects camera motions using the learned action values.
Let $\mathcal{A}_t \subseteq \mathcal{A}_{\mathrm{cam}}$ denote the motions not yet observed to collide at the current pose.
When continuing the search, the agent selects
\begin{equation}
a_t = \arg\max_{a \in \mathcal{A}_t} \hat{Q}_\theta(q,H_t,a).
\label{eq:argmax}
\end{equation}
The agent executes only the selected motion, updates its observation history, and recomputes the action values at the next step.

A colliding motion leaves the pose unchanged while consuming a step (Sec.~\ref{sec:problem}), so the policy excludes it from $\mathcal{A}_t$ until the camera pose changes, using only observed outcomes and no privileged scene geometry.

The learned LoRA adapters are enabled only for action-value prediction.
Stopping uses the base VLM with these adapters disabled.
Candidate scoring requires no rendering or generation of future views.

\section{Experiments}
\label{sec:experiments}

\subsection{Experimental Setup}
\label{sec:setup}
We ask four questions: (i) whether ranking camera motions by predicted answerability improves viewpoint selection over direct control, action imitation, and future-view generation (Sec.~\ref{sec:main-results}); (ii) whether answerability is a reliable supervision signal (Sec.~\ref{sec:exp-signal}); (iii) which design and data choices drive the gain (Sec.~\ref{sec:ablation}); and (iv) whether the simulation-trained policy transfers to a physical robot (Sec.~\ref{sec:realworld}).

\textbf{Dataset.}
We use E3VS-Bench~\cite{sakamoto2026e3vsbench} because its questions require camera motions that reveal viewpoint-dependent evidence, and its 3DGS scenes allow rendering arbitrary viewpoints for the privileged teacher.
It comprises 2{,}014 episodes across six question types in 99 indoor 3DGS scenes from SceneSplat++~\cite{ma2025scenesplat++}.
Each episode specifies only an initial pose and a human-annotated goal viewpoint, so training trajectories must be generated by the teacher (Sec.~\ref{sec:teacher}).
The official scene-disjoint split has 1{,}406 training (68 scenes), 231 validation (10 scenes), and 377 test episodes (21 unseen scenes).

\textbf{Metrics.}
GPT-5.1~\cite{openai2025gpt51} answers each question from the final view, and a GPT-5.5~\cite{openai2026gpt55} judge scores the answer from 1 to 5.
The primary metric is the mean judge score (Avg.), reported overall and by question type; an episode is successful if it scores 5.
We also report the mean episode length (Steps), the collision rate (Coll., the fraction of episodes with at least one collision), and the repetition rate (Rep.), which captures oscillatory motion: for executed camera motions $a_1,\dots,a_N$ with $N \ge 3$, $\mathrm{Rep} = \frac{1}{N-2}\sum_{t=1}^{N-2}\mathbf{1}[a_t = a_{t+2}]$, averaged over episodes.

\textbf{Baselines.}
All methods use the same episodes, action budget, final-view VQA model, and judge.
We compare direct camera control using GPT-5.1~\cite{openai2025gpt51}, Gemini 3.0 Flash~\cite{gemini3blog}, and Qwen3.6-27B~\cite{qwen2026qwen36}.
These agents receive only the current view, as the benchmark found that additional views do not change their scores~\cite{sakamoto2026e3vsbench}.
To compare value supervision with action imitation, we also fine-tune Qwen3.6-27B and $\pi_{0.5}$~\cite{black2025pi05} to clone the selected actions of the same teacher trajectories.
As a baseline that predicts future views at deployment, Cosmos3 (Planning) generates each candidate's future view with Cosmos3-Nano~\cite{nvidia2026cosmos3} fine-tuned on the same teacher trajectories and scores it with the zero-shot probe.

\textbf{Implementation.}
Observations are $512\times512$ RGB, $T_{\max}=25$, and camera motions use 0.25\,m translation and $30^{\circ}$ rotation increments.
Both the teacher and the student use Qwen3.6-27B~\cite{qwen2026qwen36}.
Rank-16 LoRA adapters ($\alpha=32$, w/o dropout) on all language-model linear projections give 117M trainable parameters; the vision encoder is frozen.
We optimize Eq.~(\ref{eq:loss}) on approximately 132k teacher-labeled state--action pairs with AdamW (constant learning rate $10^{-4}$, gradient-norm clipping 1.0).
Training runs data-parallel on two RTX PRO 6000 Blackwell GPUs with bfloat16 weights and gradient checkpointing, one state (about ten candidate probes) per GPU per step.
Stage 1 trains on $Q_1$ for up to three epochs and stage 2, resuming from the selected stage-1 checkpoint, on $Q_2$ for up to two; each stage keeps the per-epoch checkpoint with the lowest BCE on 200 held-out validation states.
Unless otherwise noted, results are reported on the test set.

\begin{table}[t]
\centering
\caption{\textbf{Main results on the E3VS-Bench test set.} Green is best and light green is second best per column. OS = object search, OST = object state, OA = object attribute, CGS = context-guided search, SR = spatial reasoning, CNT = counting.}
\label{tab:main}
\setlength{\tabcolsep}{2pt}
\begin{tabular}{lcccccc@{\hspace{6pt}}cc}
\toprule
Method & OS & OST & OA & CGS & SR & CNT & Avg. $\uparrow$ & Coll. $\downarrow$ \\
\midrule
\multicolumn{9}{l}{\emph{Proprietary}} \\
GPT-5.1 & \secondc3.34 & \secondc2.73 & \firstc2.96 & 2.87 & 2.38 & \secondc2.28 & \secondc2.81 & 0.38 \\
Gemini 3.0 Flash & 3.14 & \firstc3.09 & 2.67 & 2.47 & 2.25 & 1.96 & 2.62 & 0.43 \\
\midrule
\multicolumn{9}{l}{\emph{Open-source}} \\
Cosmos3 (Planning) & 2.83 & 2.18 & 2.38 & 2.33 & 2.25 & 1.92 & 2.36 & 0.50 \\
Qwen3.6-27B & 2.79 & 1.82 & 2.09 & 2.47 & 1.50 & 1.64 & 2.14 & 0.55 \\
$\pi_{0.5}$ & 2.59 & 2.18 & 2.24 & 1.93 & 1.75 & 1.76 & 2.15 & \secondc0.19 \\
\midrule
\textbf{\method{} (Ours)} & \firstc3.55 & 2.55 & \secondc2.89 & \firstc3.67 & \firstc2.88 & \firstc2.64 & \firstc3.05 & \firstc0.11 \\
\bottomrule
\vspace{-2.0em}
\end{tabular}
\end{table}

\subsection{Main Results}
\label{sec:main-results}

Table~\ref{tab:main} compares \method{} with direct action prediction, action imitation, and future-view generation.
\method{} achieves a mean judge score of 3.05 against 2.14 for direct action prediction with the same base VLM and stopping mechanism, so the learned viewpoint-selection policy accounts for a large part of the gain.
Cosmos3 (Planning), which generates candidate future views, scores 2.36, and GPT-5.1, the strongest proprietary baseline evaluated, 2.81.

\method{} is highest on four of the six question types, while GPT-5.1 leads OA (2.96 vs.\ 2.89) and Gemini 3.0 Flash leads OST (3.09 vs.\ 2.55), discussed in Sec.~\ref{sec:discussion}.

\begin{figure}[t]
    \centering
    \includegraphics[width=0.8\linewidth]{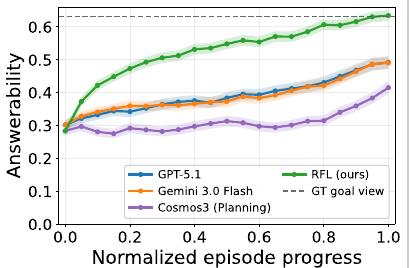}
    \caption{\textbf{Answerability along test rollouts} for the agents of Table~\ref{tab:main} (bands: $\pm 1$\,SE; dashed: mean at goal views).}
    \label{fig:transition}
    \vspace{-0.75em}
\end{figure}

\begin{figure*}[t]
    \centering
    \begin{minipage}[t]{0.32\linewidth}
        \centering
        \includegraphics[width=\linewidth]{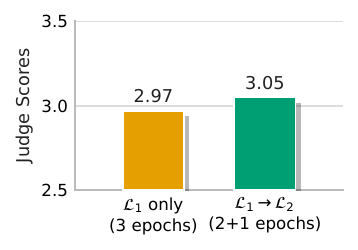}
        \vspace{-2.2em}
        \caption{\textbf{Two-step supervision} at an equal three-epoch budget (test split).}
        \label{fig:stage}
    \end{minipage}\hfill
    \begin{minipage}[t]{0.32\linewidth}
        \centering
        \includegraphics[width=\linewidth]{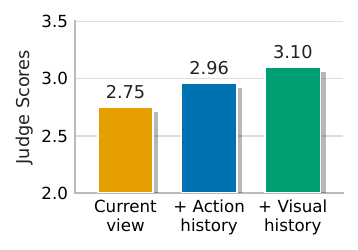}
        \vspace{-2.2em}
        \caption{\textbf{Probe conditioning} with the same two-stage recipe (validation split).}
        \label{fig:conditioning}
    \end{minipage}\hfill
    \begin{minipage}[t]{0.32\linewidth}
        \centering
        \includegraphics[width=\linewidth]{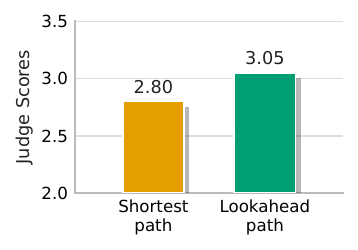}
        \vspace{-2.2em}
        \caption{\textbf{Effect of teacher trajectories} (test split).}
        \label{fig:trajectories}
    \end{minipage}
    \vspace{-1.2em}
\end{figure*}
%
%
\begin{figure*}[t]
    \centering
    \includegraphics[width=\linewidth]{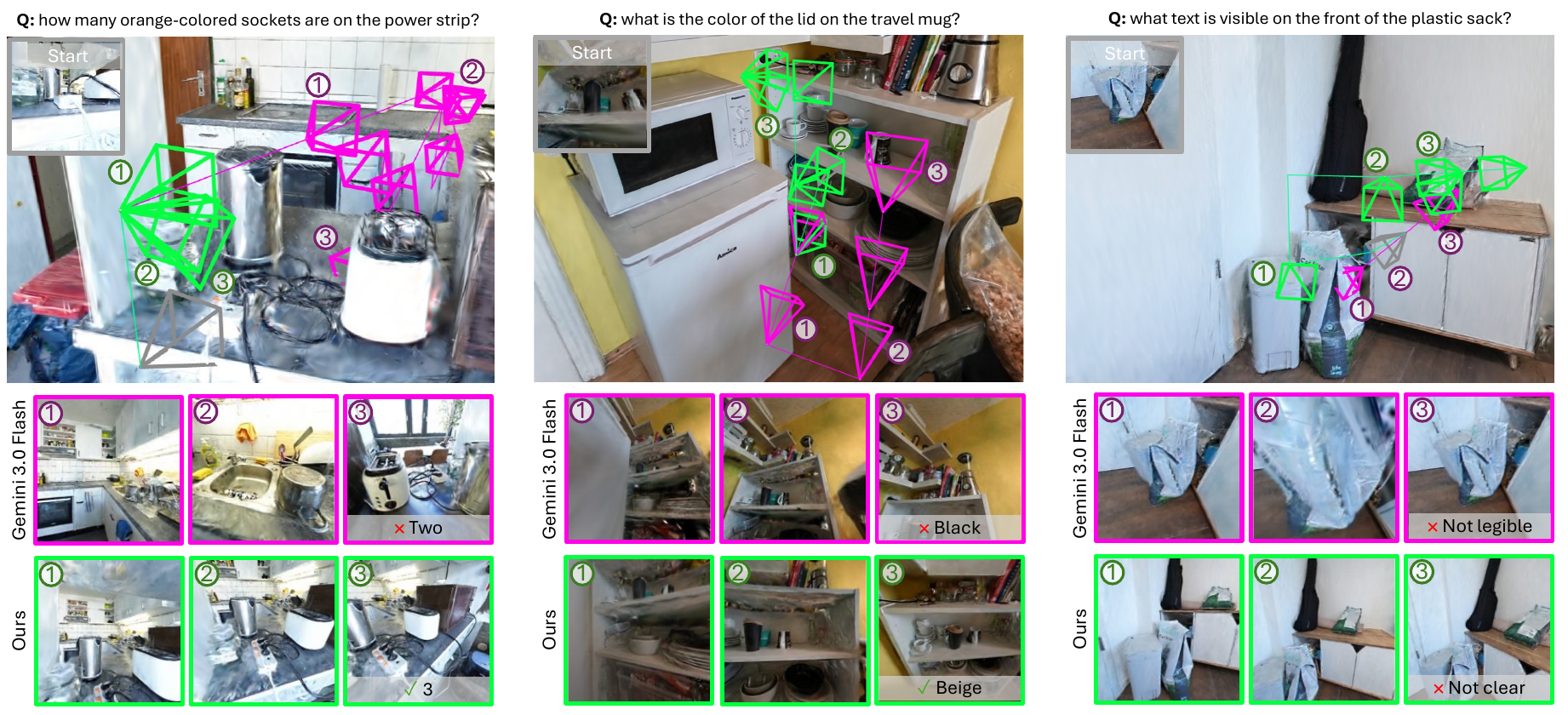}
    \vspace{-1.6em}
    \caption{Qualitative comparison on E3VS-Bench (\method{}: green; Gemini 3.0 Flash: magenta). Full answers (abbreviated in the figure): middle, \method{}: ``The lid on the travel mug is beige.''; right, Gemini: ``The text on the front of the plastic sack is not legible.'', \method{}: ``The text is not clear enough to be read.''}
    \label{fig:qualitative}
    \vspace{-2.0em}
\end{figure*}
%
%
\begin{figure*}[t]
    \centering
    \includegraphics[width=0.95\linewidth]{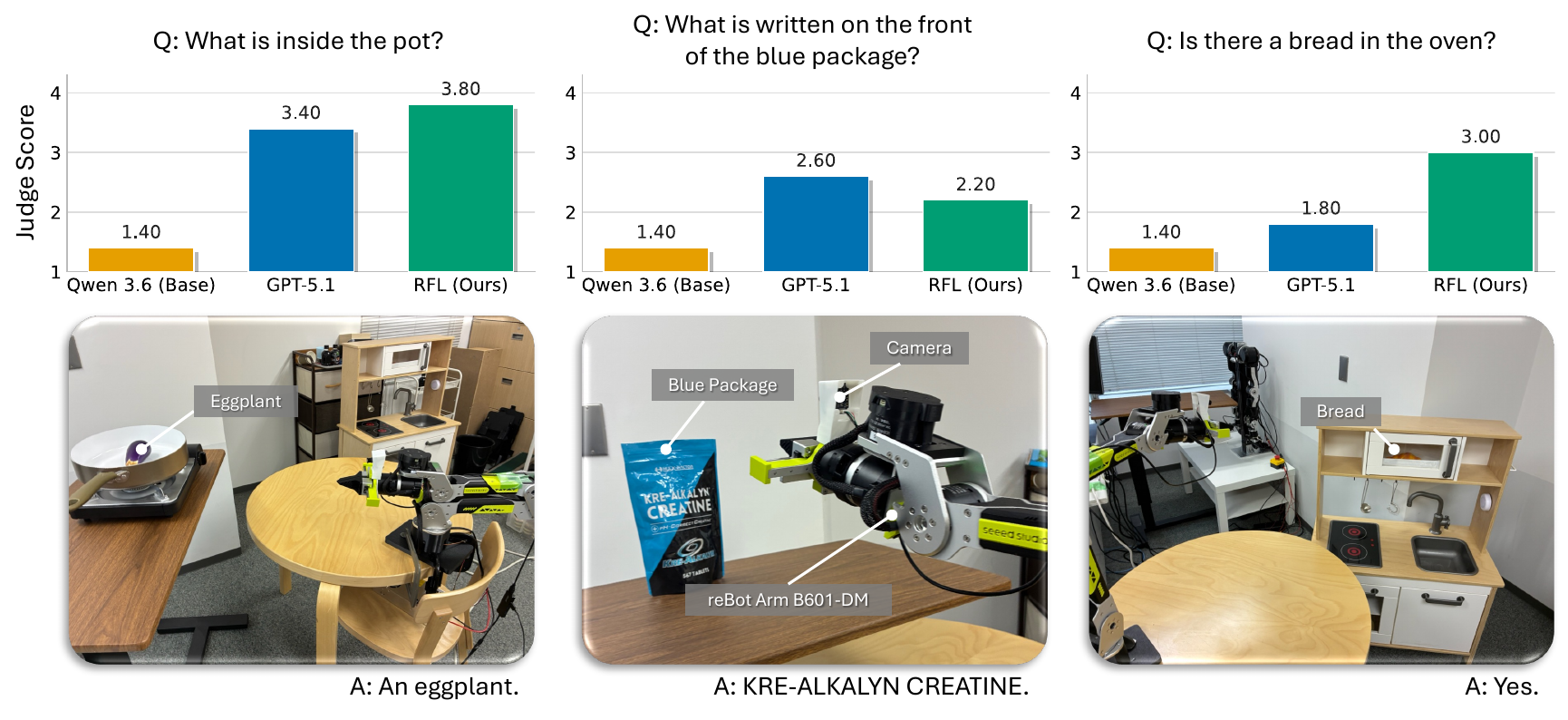}
    \vspace{-0.8em}
    \caption{\textbf{Real-world experiments.} Mean judge score per question for \method{}, the direct base VLM (Base), and the GPT-5.1 agent. Each bar averages ten trials of one method on one question, i.e., 3 questions $\times$ 3 methods $\times$ 10 trials $= 90$ trials.}
    \label{fig:real}
    \vspace{-1.35em}
\end{figure*}

To trace the source of this gain, Fig.~\ref{fig:transition} probes every view visited on the test set with the frozen probe (Eq.~\ref{eq:probe}).
The dashed line marks the mean answerability at the human-annotated goal views, where each question is answerable.
\method{} raises answerability from the first steps and reaches the goal-view level by episode end, whereas GPT-5.1 and Gemini 3.0 Flash climb slowly to well below it and Cosmos3 (Planning) stays near its initial value for most of the episode.
\method{} thus steadily moves toward viewpoints from which the question becomes answerable, which underlies its higher scores in Table~\ref{tab:main}.

\subsection{Answerability as a Supervision Signal}
\label{sec:exp-signal}
\method{} is supervised entirely by probed answerability, so we verify that this signal tracks search success before attributing the gains to it (Fig.~\ref{fig:prelim}).
We apply the frozen Qwen3.6-27B probe~\cite{qwen2026qwen36} (Eq.~(\ref{eq:probe})) to every visited view in Gemini 3.0 Flash rollouts~\cite{gemini3blog} on the 231 validation episodes (90 successes, 141 failures), disjoint from the test set.
Mean answerability is $0.334$ at initial viewpoints and $0.685$ at ground-truth goal viewpoints.
Over normalized episode progress, it rises toward the goal-view level in successful episodes but drops early and stays low in failed ones.
Answerability distinguishes successful from failed episodes with an AUROC of $0.824$, versus $0.719$ for privileged target-object IoU and $0.743$ for episode length.
These results support using view-level answerability to construct the teacher's lookahead targets.

\subsection{Ablation Studies}
\label{sec:ablation}
We isolate four design and data choices: what the model outputs, how far the targets look ahead, what the probe observes, and which trajectories supply the training states.
All model and hyperparameter choices were made on the validation split; test-set ablations are post-hoc analyses not used for model selection.

\begin{table}[t]
\centering
\caption{\textbf{Output and fine-tuning of the same VLM} (Qwen3.6-27B) with single-view conditioning.}
\label{tab:ft}
\setlength{\tabcolsep}{8pt} 
\begin{tabular}{lcccc}
\toprule
Output & FT & Avg. $\uparrow$ & Steps $\downarrow$ & Rep. $\downarrow$ \\
\midrule
\multirow{2}{*}{Actions} & & 2.14 & 20.1 & 0.78 \\
 & \checkmark & 2.47 & 24.9 & 0.87 \\
\midrule
\multirow{2}{*}{Answerability} & & 2.04 & 18.8 & 0.68 \\
 & \checkmark & \textbf{2.71} &	\textbf{17.2} & \textbf{0.62} \\
\bottomrule
\end{tabular}
\vspace{-1.55em}
\end{table}

\textbf{Action-value supervision.}
Table~\ref{tab:ft} compares direct action prediction and answerability-based action scoring, with and without fine-tuning, on the same Qwen3.6-27B backbone.
All four variants observe only the current view, and both fine-tuned models use the same teacher trajectories: action prediction imitates the selected motions, whereas value prediction learns answerability targets for all candidate motions.
The latter is the single-view variant of \method{}; its full-history counterpart is the deployed model (3.05).

Without fine-tuning, answerability scoring trails direct action prediction (2.04 vs.\ 2.14), and fine-tuning lifts value prediction to 2.71 against 2.47 for action cloning.
The value-based policy is also shorter (17.2 vs.\ 24.9 steps) and less repetitive (0.62 vs.\ 0.87).
These results support learning candidate-wise action values rather than imitating the teacher's selected motions.

\begin{table}[t]
\centering
\caption{\textbf{Render oracles vs.\ \method{}}.}
\vspace{-0.5em}
\label{tab:oracle}
\setlength{\tabcolsep}{8pt} 
\begin{tabular}{lcccc}
\toprule
Method & Horizon & Avg. $\uparrow$ & Steps $\downarrow$ & Rep. $\downarrow$ \\
\midrule
\multirow{2}{*}{\method{}} & 1 & 2.98 & 16.0 & 0.50 \\
 & 2 & 3.05 & 16.1 & 0.46 \\
\midrule
\multirow{2}{*}{Oracle} & 1 & 2.98 & 14.5 & 0.51 \\
 & 2 & 3.19 & 13.6 & 0.37 \\
\bottomrule
\end{tabular}
\vspace{-2.0em}
\end{table}

\textbf{Lookahead horizon.}
To examine how much lookahead depth survives without rendering, Table~\ref{tab:oracle} compares one- and two-step variants of \method{} with privileged rendering oracles.
The oracles apply the teacher at inference, rendering each candidate's actual future view, scoring it with the frozen VLM, and executing the highest-scoring motion; \method{} instead predicts action values without observing future views.

Going from one to two steps raises \method{} from 2.98 to 3.05 and the oracle from 2.98 to 3.19: \method{} matches the oracle at one step and trails it by 0.14 at two.
The two-step oracle is also shorter (13.6 vs.\ 16.1 steps) and less repetitive (0.37 vs.\ 0.46).
To separate two-step supervision from additional training, we fix the total budget at three epochs (Fig.~\ref{fig:stage}).
Three epochs on $\mathcal{L}_1$ alone reach 2.97, whereas the deployed schedule of two epochs on $\mathcal{L}_1$ and one on $\mathcal{L}_2$ reaches 3.05 (the one-step entry of Table~\ref{tab:oracle}, 2.98, differs only in its validation-selected two-epoch checkpoint).
This controlled comparison indicates that the modest gain is not due to optimization alone and is consistent with a benefit from the two-step targets.
Together with the oracle's larger horizon gain, this suggests that two-step lookahead provides useful supervision whose full benefit is not yet transferred into a rendering-free value predictor.

\textbf{Observation history.}
We compare three input configurations on the validation set under the same two-stage recipe (Fig.~\ref{fig:conditioning}).
All variants receive the question and candidate motion, plus either the current view, the current view and the last three executed motions, or recent observations.
Action history raises the mean judge score from 2.75 to 2.96, and visual history to 3.10, supporting its use in the final model.

\textbf{Training trajectories.}
We compare \method{} trained on states along shortest paths, planned with A$^{*}$ to the goal viewpoint, with \method{} trained on lookahead-teacher trajectories (Fig.~\ref{fig:trajectories}).
Under the same two-stage recipe, lookahead trajectories reach 3.05 against 2.80 for shortest paths, so the states induced by the answerability-driven teacher are more effective for training than geometrically efficient paths.

\subsection{Qualitative Results}
\label{sec:qualitative}

To see how the learned value shapes search behavior, Fig.~\ref{fig:qualitative} shows three test episodes.
In the left two, \method{} moves directly to a viewpoint that reveals the queried information, descending to the power strip and closing in on the travel mug, and answers correctly, whereas Gemini 3.0 Flash never reaches such a viewpoint and answers both incorrectly.
The right episode is a failure case: \method{} reaches a view that exposes the text on the sack yet fails to stop there and ends at a viewpoint from which the text is no longer readable.
Reaching an informative viewpoint is thus not sufficient: the policy must also learn when to stop, a decision \method{} currently leaves to the base VLM.
To quantify this effect, we re-answer each episode from the visited view with the highest probed answerability, which raises the score from 3.05 to 3.16 and narrows the gap to the two-step rendering oracle (3.19, Table~\ref{tab:oracle}) from 0.14 to 0.03. A substantial part of the remaining oracle gap may therefore stem from stopping rather than viewpoint selection.

\subsection{Real-World Experiments}
\label{sec:realworld}

Finally, we test whether the simulation-trained policy transfers to physical camera control, a deployment that rendering-free inference makes practical.
The platform is a Seeed Studio reBot Arm B601-DM, an open-source 6-DoF arm with an end-effector camera, whose workspace covers the benchmark's 5-DoF viewpoint actions.
We evaluate three viewpoint-dependent questions with three methods, \method{}, the base VLM acting directly, and the GPT-5.1 agent, running ten trials for every method--question pair while varying the object placement and initial arm pose, i.e., 3 questions $\times$ 3 methods $\times$ 10 trials $= 90$ trials in total (Fig.~\ref{fig:real}).
All trials use $T_{\max}=15$, the simulation increments of 0.25\,m and $30^{\circ}$, and the same scoring protocol on the final onboard view.
\method{} achieves the highest mean judge score over the three questions, reaching 3.80 on the pot and 3.00 on the oven question, whereas the direct base VLM stays at 1.40 on all three.
GPT-5.1 reaches 3.40 on the pot question but falls to 1.80 on the oven question, and on the blue-package question, which requires reading the text on its front, it edges out \method{} (2.60 vs.\ 2.20).
These experiments provide initial evidence of such transfer without test-time rendering.
The arm's embodiment, however, limits execution: motions infeasible from the current pose are removed from \method{}'s candidates before selection, whereas the direct base VLM kept generating them even after they were excluded from its action list.
A model's intended motion can thus be unexecutable for a given embodiment, which calls for policies that account for it.

\subsection{Discussion and Limitations}
\label{sec:discussion}

\method{} requires counterfactual rendering at training time.
Any free-viewpoint renderer (mesh, NeRF, 3DGS) would serve, whereas logged trajectories alone label only executed actions, weakening the dense counterfactual supervision.
The recipe is off-policy privileged value distillation: the student trains on the teacher's visit distribution and deploys on its own, yet this covariate shift appears cheap, as \method{} matches the one-step render oracle scoring its on-policy states (Table~\ref{tab:oracle}).
The remaining weakness is fine-grained evidence: OST is \method{}'s weakest type (2.55 vs.\ 3.09 for Gemini 3.0 Flash), as it hinges on details requiring precisely placed close views.
Stopping is a further limitation: as the OA failure in Fig.~\ref{fig:qualitative} shows, \method{} can reach a view exposing the answer yet fail to stop there, and the best-visited-view analysis (Sec.~\ref{sec:qualitative}) attributes much of the oracle gap to this decision, left to the base VLM.
Real-world experiments add an embodiment limitation, as motions feasible in simulation are not always executable by the arm.

\section{Conclusion}
\label{sec:conclusion}

We presented \method{}, a rendering-free viewpoint-planning method for Embodied 3D Visual Search.
Lookahead is confined to a privileged teacher that counterfactually renders every candidate action in 3DGS training scenes and scores it by two-step answerability beam search.
The same VLM's action-conditioned Yes/No probability is then calibrated to these measured values by two-stage soft-label distillation, so the deployed policy selects viewpoints greedily with a single text probe per candidate action, without rendering, world models, or generation.
On E3VS-Bench, \method{} raised the direct baseline's mean judge score from 2.14 to 3.05 and outperformed action cloning trained on the same supervision.
Ablations support the recipe, as the zero-shot probe underperforms even the direct baseline.
In real-world experiments on a 6-DoF arm, the simulation-trained policy achieved the highest mean judge score across three viewpoint-dependent questions.

Future work includes calibrated stopping, and finer-grained or continuous action spaces for physical platforms.

\section{Acknowledgments}
\label{sec:acknowledgments}
This work was supported by JST PRESTO (Grant Number JPMJPR22P8), JST SPRING (Grant Number JPMJSP2108), JSPS KAKENHI (Grant Number 25K03177), Japan.
The first author used Claude Code (Anthropic) for manuscript drafting and code generation, and ChatGPT (OpenAI) for language editing and proofreading. AI-assisted outputs were reviewed and revised by the first author as appropriate, who takes responsibility for the final manuscript and implementation.

\bibliographystyle{splncs04}
\bibliography{main}

@String(CVPR = {CVPR})

@String(ICCV = {ICCV})

@String(ECCV = {ECCV})

@String(IROS = {IROS})

@String(RSS = {RSS})

@String(TOG = {TOG})

@String(CoRL = {CoRL})

@String(NeurIPS = {NeurIPS})

@String(ICLR = {ICLR})

@inproceedings{sakamoto2026e3vsbench,
  title     = {{E3VS-Bench}: A Benchmark for Viewpoint-Dependent Active Perception in {3D} {Gaussian} Splatting Scenes},
  author={Koya Sakamoto and Taiki Miyanishi and Daichi Azuma and Shuhei Kurita and Shu Morikuni and Naoya Chiba and Motoaki Kawanabe and Yusuke Iwasawa and Yutaka Matsuo},
  booktitle = ECCV,
  year      = {2026},
}

@inproceedings{mahumar2024openeqa,
    title = {OpenEQA: Embodied Question Answering in the Era of Foundation Models},
    booktitle=CVPR,
    author = {Majumdar, Arjun and Ajay, Anurag and Zhang, Xiaohan and Putta, Pranav and Yenamandra, Sriram and Henaff, Mikael and Silwal, Sneha and Mcvay, Paul and Maksymets, Oleksandr and Arnaud, Sergio and Yadav, Karmesh and Li, Qiyang and Newman, Ben and Sharma, Mohit and Berges, Vincent and Zhang, Shiqi and Agrawal, Pulkit and Bisk, Yonatan and Batra, Dhruv and Kalakrishnan, Mrinal and Meier, Franziska and Paxton, Chris and Sax, Sasha and Rajeswaran, Aravind},
    year={2024},
}

@misc{koo2025towardb,
  title={Toward Ambulatory Vision: Learning Visually-Grounded Active View Selection},
  author={Juil Koo and Daehyeon Choi and Sangwoo Youn and Phillip Y. Lee and Minhyuk Sung},
  year          = {2025},
  eprint        = {2512.13250},
  archivePrefix = {arXiv},
  primaryClass  = {cs.CV},
  url           = {https://arxiv.org/abs/2512.13250}
}

@misc{wang2026planning,
  title={Planning with the Views},
  author={Kangrui Wang and Linjie Li and Zhengyuan Yang and Shiqi Chen and Zihan Wang and Li Fei-Fei and Jiajun Wu and Leonidas Guibas and Lijuan Wang and Manling Li},
  year={2026},
  eprint={2605.29563},
  archivePrefix={arXiv},
  primaryClass={cs.AI},
  url={https://arxiv.org/abs/2605.29563}
}

@inproceedings{ren2024explore,
  title     = {Explore until Confident: Efficient Exploration for Embodied Question Answering},
  author    = {Ren, Allen Z. and Clark, Jaden and Dixit, Anushri and Itkina, Masha and Majumdar, Anirudha and Sadigh, Dorsa},
  booktitle = RSS,
  year      = {2024}
}

@inproceedings{azuma2024answerability,
  title={Answerability Fields: Answerable Location Estimation via Diffusion Models}, 
  author={Daichi Azuma and Taiki Miyanishi and Shuhei Kurita and Koya Sakamoto and Motoaki Kawanabe},
  booktitle=IROS,
  year={2024},
}

@article{kerbl2023GaussianSplatting,
    title={{3D} {G}aussian splatting for real-time radiance field rendering},
    author = {Kerbl, Bernhard and Kopanas, Georgios and Leimk{\"u}hler, Thomas and Drettakis, George},
    journal=TOG,
    volume={42},
    number={4},
    pages={1--14},
    month={July},
    year={2023},
    publisher={ACM New York, NY, USA}
}

@article{Bajcsy1988ActivePerception,
  author = {Ruzena Bajcsy},
  title = {Active Perception},
  journal = {Proceedings of the IEEE},
  year = {1988},
  volume = {76},
  number = {8},
  pages = {966--1005}
}

@article{Aloimonos1988ActiveVision,
  author = {John Aloimonos and Isaac Weiss and Amit Bandyopadhyay},
  title = {Active Vision},
  journal = {International Journal of Computer Vision},
  year = {1988},
  volume = {1},
  pages = {333--356}
}

@inproceedings{kerrj2025eyerobot,
  title={Eye, Robot: Learning to Look to Act with a BC-RL Perception-Action Loop},
  author={Justin Kerr and Kush Hari and Ethan Weber and Chung Min Kim and Brent Yi and Tyler Bonnen and Ken Goldberg and Angjoo Kanazawa},
  booktitle=CORL,
  year={2025},
}

@inproceedings{das2018embodied,
  title={{E}mbodied {Q}uestion {A}nswering},
  author={Abhishek Das and Samyak Datta and Georgia Gkioxari and Stefan Lee and Devi Parikh and Dhruv Batra},
  booktitle = CVPR,
  year      = {2018}
}

@inproceedings{grapheqa2025,
    title={GraphEQA: Using 3D Semantic Scene Graphs for Real-time Embodied Question Answering},
    author={Saxena, Saumya and Buchanan, Blake and Paxton, Chris and Liu, Peiqi and Chen, Bingqing and Vaskevicius, Narunas and Palmieri, Luigi and Francis, Jonathan and Kroemer, Oliver},
    booktitle=CoRL,
    year={2025}
}

@inproceedings{sakamoto2024mapeqa,
  author={Koya Sakamoto and Daichi Azuma and Taiki Miyanishi and Shuhei Kurita and Motoaki Kawanabe},
  title={Map-based Modular Approach for Zero-shot Embodied Question Answering},
  booktitle=IROS,
  year={2024},
}

@inproceedings{koh2021pathdreamer,
  title={Pathdreamer: A World Model for Indoor Navigation}, 
  author={Koh, Jing Yu and Lee, Honglak and Yang, Yinfei and Baldridge, Jason and Anderson, Peter},
  booktitle = ICCV,
  year      = {2021}
}

@inproceedings{bar2025navigation,
  title     = {Navigation World Models},
  author={Amir Bar and Gaoyue Zhou and Danny Tran and Trevor Darrell and Yann LeCun},
  booktitle = CVPR,
  year      = {2025}
}

@misc{azuma2026navwam,
  title={NavWAM: A Navigation World Action Model for Goal-Conditioned Visual Navigation}, 
  author={Daichi Azuma and Taiki Miyanishi and Koya Sakamoto and Shuhei Kurita and Yaonan Zhu and Petr Khrapchenkov and Motoaki Kawanabe and Yusuke Iwasawa and Yutaka Matsuo},
  year          = {2026},
  eprint        = {2606.13494},
  archivePrefix = {arXiv},
  primaryClass  = {cs.RO},
  url           = {https://arxiv.org/abs/2606.13494}
}

@inproceedings{wang2024lookahead,
  title     = {Lookahead Exploration with Neural Radiance Representation for Continuous Vision-Language Navigation},
  author    = {Wang, Zihan and Li, Xiangyang and Yang, Jiahao and Liu, Yeqi and Hu, Junjie and Jiang, Ming and Jiang, Shuqiang},
  booktitle = CVPR,
  year      = {2024}
}

@misc{lu2026gsmem,
  title={GSMem: 3D Gaussian Splatting as Persistent Spatial Memory for Zero-Shot Embodied Exploration and Reasoning}, 
  author={Yiren Lu and Yi Du and Disheng Liu and Yunlai Zhou and Chen Wang and Yu Yin},
  year          = {2026},
  eprint        = {2603.19137},
  archivePrefix = {arXiv},
  primaryClass  = {cs.CV},
  url           = {https://arxiv.org/abs/2603.19137}
}

@misc{yuan2026fastwam,
  title         = {Fast-WAM: Do World Action Models Need Test-time Future Imagination?},
  author        = {Tianyuan Yuan and Zibin Dong and Yicheng Liu and Hang Zhao},
  year          = {2026},
  eprint        = {2603.16666},
  archivePrefix = {arXiv},
  primaryClass  = {cs.CV},
  url={https://arxiv.org/abs/2603.16666}
}

@misc{chen2026worldmap,
  title={WorldMAP: Bootstrapping Vision-Language Navigation Trajectory Prediction with Generative World Models}, 
  author={Hongjin Chen and Shangyun Jiang and Tonghua Su and Chen Gao and Xinlei Chen and Yong Li and Zhibo Chen},
  year          = {2026},
  eprint        = {2604.07957},
  archivePrefix = {arXiv},
  url           = {https://arxiv.org/abs/2604.07957}
}

@inproceedings{chen2019learning,
  title     = {Learning by Cheating},
  author    = {Chen, Dian and Zhou, Brady and Koltun, Vladlen and Kr{\"a}henb{\"u}hl, Philipp},
  booktitle = CoRL,
  pages     = {66--75},
  year      = {2019}
}

@article{lee2020learning,
  author  = {Lee, Joonho and Hwangbo, Jemin and Wellhausen, Lorenz and Koltun, Vladlen and Hutter, Marco},
  title   = {Learning Quadrupedal Locomotion over Challenging Terrain},
  journal = {Science Robotics},
  volume  = {5},
  number  = {47},
  pages   = {eabc5986},
  year    = {2020},
  doi     = {10.1126/scirobotics.abc5986},
}

@inproceedings{ramakrishnan2018sidekick,
  title     = {Sidekick Policy Learning for Active Visual Exploration},
  author    = {Ramakrishnan, Santhosh K. and Grauman, Kristen},
  booktitle = ECCV,
  year      = {2018}
}

@inproceedings{rusu2016policy,
  title     = {Policy Distillation},
  author={Andrei A. Rusu and Sergio Gomez Colmenarejo and Caglar Gulcehre and Guillaume Desjardins and James Kirkpatrick and Razvan Pascanu and Volodymyr Mnih and Koray Kavukcuoglu and Raia Hadsell},
  booktitle = ICLR,
  year      = {2016}
}

@inproceedings{ruoss2024amortized,
  title     = {Amortized Planning with Large-Scale Transformers: A Case Study on Chess},
  author = {Ruoss, Anian and Del\'{e}tang, Gr\'{e}goire and Medapati, Sourabh and Grau-Moya, Jordi and Wenliang, Li Kevin and Catt, Elliot and Reid, John and Lewis, Cannada A. and Veness, Joel and Genewein, Tim},
  booktitle = NeurIPS,
  year      = {2024}
}

@inproceedings{hong2025qsft,
  title = {Q-SFT: Q-Learning for Language Models via Supervised Fine-Tuning},
  author    = {Hong, Joey and Dragan, Anca and Levine, Sergey},
  booktitle = ICLR,
  year      = {2025}
}

@inproceedings{bai2025digiq,
  title={Digi-Q: Transforming VLMs to Device-Control Agents via Value-Based Offline RL},
    author={Bai, Hao and Zhou, Yifei and Li, Erran Li and Levine, Sergey and Kumar, Aviral},
  booktitle = ICLR,
  year      = {2025}
}

@inproceedings{ma2025scenesplat++,
  title     = {SceneSplat++: A Large Dataset and Comprehensive Benchmark for Language Gaussian Splatting},
  author    = {Ma, Mengjiao and Ma, Qi and Li, Yue and Cheng, Jiahuan and Yang, Runyi and Ren, Bin and Popovic, Nikola and Wei, Mingqiang and Sebe, Nicu and Van Gool, Luc and Gevers, Theo and Oswald, Martin R. and Paudel, Danda Pani},
  booktitle=NeurIPS,
year={2025}
}

@misc{openai2025gpt51,
  author       = {{OpenAI}},
  title        = {GPT-5.1: A smarter, more conversational ChatGPT},
  howpublished = {\url{https://openai.com/index/gpt-5-1/}},
  year         = {2025},
  month        = {11},
  note         = {Accessed: 2026-09-15}
}

@misc{openai2026gpt55,
  author       = {{OpenAI}},
  title        = {Introducing GPT-5.5},
  howpublished = {\url{https://openai.com/index/introducing-gpt-5-5/}},
  year         = {2026},
  month        = {4},
  note         = {Accessed: 2026-08-19}
}

@misc{gemini3blog,
  author       = {{Google DeepMind}},
  title        = {A new era of intelligence with Gemini 3},
  year         = {2025},
  month        = nov,
  howpublished = {\url{https://blog.google/products/gemini/gemini-3/}},
  note         = {Accessed: 2026-09-13}
}

@misc{qwen2026qwen36,
  author       = {{Qwen Team}},
  title        = {Qwen3.6-27B: Flagship-Level Coding in a 27B Dense Model},
  howpublished = {\url{https://qwen.ai/blog?id=qwen3.6-27b}},
  year         = {2026},
  month        = {4},
  note         = {Accessed: 2026-08-19}
}

@misc{black2025pi05,
  title         = {$\pi_{0.5}$: a Vision-Language-Action Model with Open-World Generalization},
  author        = {{Physical Intelligence} and Kevin Black and Noah Brown and James Darpinian and Karan Dhabalia and Danny Driess and Adnan Esmail and Michael Equi and Chelsea Finn and Niccolo Fusai and others},
  year          = {2025},
  eprint        = {2504.16054},
  archivePrefix = {arXiv},
  url           = {https://arxiv.org/abs/2504.16054}
}

@misc{nvidia2026cosmos3,
  author  = {{NVIDIA}},
  title   = {Cosmos 3: Omnimodal World Models for Physical {AI}},
  year={2026},
  eprint={2606.02800},
  archivePrefix={arXiv},
  primaryClass={cs.CV},
  url={https://arxiv.org/abs/2606.02800}, 
}


\end{document}